\documentclass[11pt]{article}
\usepackage{acl2014}
\usepackage{times}
\usepackage{url}
\usepackage{latexsym}

\usepackage{graphicx} 
\usepackage{multirow}

\title{DepecheMood: \\ a Lexicon for Emotion Analysis from Crowd-Annotated News}

  \author{Jacopo Staiano \\
University of Trento\\
Trento - Italy\\
{\tt staiano@disi.unitn.it} \\\And
Marco Guerini \\
Trento RISE\\
Trento - Italy\\
{\tt marco.guerini@trentorise.eu}
}

\date{}

\begin{document}
\maketitle
\begin{abstract}
While many lexica annotated with words polarity are available for sentiment analysis, very few tackle the harder task of emotion analysis and are
usually quite limited in coverage. In this paper, we present a novel approach for extracting -- in a totally automated way -- a high-coverage and high-precision lexicon of roughly 37 thousand terms annotated  
with emotion scores, called \texttt{DepecheMood}. 
Our approach exploits in an original way `crowd-sourced' affective annotation implicitly provided by readers of news articles from \texttt{rappler.com}. 
By providing new state-of-the-art performances in unsupervised settings for regression and classification tasks, even using a na\"{\i}ve approach, our experiments show the 
beneficial impact of harvesting
social media data for affective lexicon building.

\end{abstract}

\section{Introduction}

Sentiment analysis has proved useful in several application scenarios,
for instance in buzz monitoring -- the marketing technique for keeping track of consumer responses to services and 
products -- where identifying positive and negative customer experiences helps to assess product and service demand, tackle crisis management, etc.


On the other hand, the use of finer-grained models, accounting for the role of individual emotions, is still in its infancy. 
The simple division in 
`positive' vs. `negative' comments may not suffice, as in these examples: `\emph{I'm so miserable, I dropped my IPhone in the water and now it's not working anymore}' 
(\textsc{sadness}) vs. `\emph{I am very upset, my new IPhone keeps not working!}' (\textsc{anger}). While both texts express a negative sentiment, the latter, connected to anger, is more relevant 
for buzz monitoring. 
Thus, emotion analysis represents a natural evolution of sentiment analysis. 

Many approaches to sentiment analysis make use of lexical resources -- i.e. lists of positive and negative words -- often deployed as baselines or as features for other methods, 
usually machine learning based~\cite{liu2012survey}. In these lexica, words are associated with their prior polarity, i.e. whether such word out of context evokes something positive or 
something negative. For example, \emph{wonderful} has a positive connotation -- prior polarity -- while \emph{horrible} has a negative one. 

The quest for a high precision and high coverage lexicon, where words are associated with either sentiment or emotion scores, 
has several reasons.
First,
it is fundamental for tasks such as affective modification of existing texts, where words' polarity together with 
their score are necessary for creating multiple \emph{graded} variations of the original text~\cite{inkpen2006generating,Guerini2008,Whitehead10}. 

Second, considering word order makes a difference in sentiment analysis. This calls 
for a role of compositionality, where the score of a sentence is computed by composing the scores of the words up in the syntactic tree.  Works worth mentioning in this 
connection are: \newcite{socher2013recursive}, which uses recursive neural networks to learn compositional rules for sentiment analysis, and \cite{neviarouskaya2009compositionality,neviarouskaya2011affect} 
which exploit hand-coded rules to compose the emotions expressed by words in a sentence. 
In this respect, compositional approaches represent a new promising trend, since all other approaches, either using semantic similarity or Bag-of-Words (BOW) based machine-learning, cannot 
handle, for example, cases of texts with same wording but different words order:
``\emph{The dangerous killer escaped one month ago, but recently he was arrested}" (\textsc{relief}, \textsc{happyness}) vs. ``\emph{The dangerous killer was arrested one month ago, but recently he 
escaped}" (\textsc{fear}). The work in ~\cite{wangbaselines} partially accounts for this problem and argues that using word bigram features allows improving over BOW based methods, where words are taken as features in isolation. This way it is possible to capture simple 
compositional 
phenomena like polarity reversing in ``\emph{killing cancer}". 

Finally, tasks such as copywriting, where evocative names are a key element to a successful product \cite{ozbalcomputational,ozbal2012brand} require exhaustive lists of emotion 
related words. In such cases no context is given and the brand name alone, with its perceived prior polarity, is responsible for stating the area of competition and evoking 
semantic associations. For example \emph{Mitsubishi} changed the name of one of its SUVs for the Spanish market, since the original name \emph{Pajero} had a very negative prior 
polarity, as it means `wanker' in Spanish \cite{piller200310}. Evoking emotions is also fundamental for a successful name: consider names of a perfume like 
\emph{Obsession}, or technological products like MacBook \emph{air}.

In this work, we aim at automatically producing a high coverage and high precision emotion lexicon using distributional semantics, with numerical scores associated with each emotion, like it has already been done for sentiment 
analysis. To this end, we take advantage in an original way of massive crowd-sourced affective annotations associated with news articles, obtained by crawling the \texttt{rappler.com} social
news network. We also evaluate our lexicon by integrating it in unsupervised classification and 
regression settings for emotion recognition. Results indicate that the use of our resource, even if automatically acquired, is highly beneficial in affective text recognition. 


\section{Related Work}
Within the broad field of sentiment analysis, we hereby provide a short 
review of research efforts put towards building sentiment and emotion lexica, regardless of the approach in which 
such lists are then used (machine learning, rule based or deep learning). A general overview can be found in~\cite{pang2008opinion,liu2012survey,wilson:AAAI-04,Paltoglou2010}.

\medskip

\begin{table*} [!htb] 	
	\begin{center} 	 	
		{\footnotesize 		
			\begin{tabular}{l|rrrrrrrrr} 		
				\hline 
 & {\scriptsize AFRAID} & {\scriptsize AMUSED} & {\scriptsize ANGRY} & {\scriptsize ANNOYED} & {\scriptsize DONT\_CARE} & {\scriptsize HAPPY} & {\scriptsize INSPIRED} & 
{\scriptsize SAD} \\
doc\_10002 & 0.75 & 0.00 & 0.00 & 0.00 & 0.00 & 0.00 & 0.25 & 0.00 \\
doc\_10003 & 0.00 & 0.50 & 0.00 & 0.16 & 0.17 & 0.17 & 0.00 & 0.00 \\
doc\_10004 & 0.52 & 0.02 & 0.03 & 0.02 & 0.02 & 0.06 & 0.02 & 0.31 \\
doc\_10011 & 0.40 & 0.00 & 0.00 & 0.20 & 0.00 & 0.20 & 0.20 & 0.00 \\
doc\_10028 & 0.00 & 0.30 & 0.08 & 0.00 & 0.00 & 0.23 & 0.31 & 0.08 \\
\hline 		
			\end{tabular} 		
		} 		 	
	\end{center}	 	
	\setlength{\belowcaptionskip}{-0.1cm} 	
	\caption{An excerpt of the Document-by-Emotion Matrix - $M_{DE}$} 	
	\label{tab:Document-by-Emotion-Matrix} 
\end{table*}  
\textbf{Sentiment Lexica}.
In recent years there has been an increasing focus on producing lists of words (lexica) with prior 
polarities, to be used in sentiment analysis. 
When building such lists, a trade-off between coverage of the resource and its precision is to be found.


One of the most well-known resources is~\emph{SentiWordNet} (SWN)~\cite{Esuli06,baccianella2010sentiwordnet}, in which each entry is associated with the numerical scores \texttt{Pos(s)} and \texttt{Neg(s)}, ranging from 0 to 1. These scores -- 
automatically assigned starting from a bunch of seed terms -- represent the positive and negative valence (or posterior polarity) of  each entry, that takes the form \texttt{lemma\#pos\#sense-number}.
Starting from SWN, several prior polarities for words (\emph{SWN-prior}), in the form \texttt{lemma\#PoS}, can be computed (e.g. considering only the first-sense, averaging on all the senses, etc.). 
These approaches, detailed in \cite{guerini2013sentiment}, produce a list of 155k words, where the lower precision given by the automatic scoring  of SWN is compensated 
by the high coverage. 

Another widely used resource is \emph{ANEW} \cite{bradley1999affective}, providing
valence scores for 1k words, which were manually assigned by several annotators. This resource has a low coverage, but the precision is maximized. 
Similarly, the \emph{SO-CAL} entries \cite{taboada2011lexicon} were
manually tagged by a small number
of annotators with a multi-class label (from \texttt{very\_negative} to \texttt{very\_positive}). These ratings were further
validated through crowd-sourcing, ending up with a list of roughly 4k words. 
More recently, a resource that replicated ANEW annotation approach using crowd-sourcing, was released~\cite{warriner2013norms}, providing sentiment scores for 14k words.
Interestingly, this resource annotates the most frequent words in English, so, even if lexicon coverage is still far lower than SWN-prior,  it grants a high coverage, with human precision, of language use. 

Finally,  the \emph{General
Inquirer} lexicon \cite{stone1966general} provides a binary
classification  (\texttt{positive}/\texttt{negative}) of 4k
sentiment-bearing words, while the resource in
\cite{wilson2005recognizing} expands the General
Inquirer to 6k words. 

\textbf{Emotion Lexica}.
Compared to sentiment lexica, far less emotion lexica have been produced, and all have lower coverage. 
One of the most used resources is \emph{WordNetAffect}~\cite{strappaLREC04} which contains manually assigned affective labels to WordNet synsets 
(\textsc{anger}, \textsc{joy}, \textsc{fear}, etc.). It currently provides 900 annotated synsets and 1.6k words in the form \texttt{lemma\#PoS\#sense}, corresponding to roughly 1 thousand 
\texttt{lemma\#PoS}. 

\emph{AffectNet}, part of the SenticNet project \cite{cambria2012sentic}, contains 10k words (out of 23k entries) taken from ConceptNet and aligned with WordNetAffect. This 
resource extends WordNetAffect labels to concepts like `have breakfast'.
\emph{Fuzzy Affect Lexicon} \cite{subasic2001affect} contains roughly 4k \texttt{lemma\#PoS} manually annotated by one linguist using 80 emotion labels.
\emph{EmoLex} \cite{mohammad2013crowdsourcing} contains almost 10k lemmas annotated with an intensity label for each emotion using Mechanical Turk.
Finally~\emph{Affect database} is an extension of SentiFul \cite{Neviarouskaya:2007fk} and contains 2.5K words in the form \texttt{lemma\#PoS}. The latter is the only lexicon 
providing words annotated also with emotion scores rather than only with labels.

\section{Dataset Collection}
\label{DS}
To build our emotion lexicon we harvested all the news articles from \texttt{rappler.com}, as of June 3rd 2013: the final dataset consists of  
13.5 M words over 25.3 K documents, with an average of 530 words per document. For each document, along with the text we also harvested the information displayed by Rappler's
\emph{Mood Meter}, a small interface 
offering the readers the opportunity to click on the emotion that a given Rappler story made them feel. The 
idea behind the Mood Meter is actually ``getting people to \emph{crowdsource} the mood for the 
day"\footnote{http://nie.mn/QuD17Z}, and returning the percentage of votes for each 
emotion label for a given story. This way, hundreds of thousands votes have been collected since the launch of the service.
 In our novel approach to `crowdsourcing', as compared to other NLP tasks that rely on tools like Amazon's Mechanical Turk \cite{snow2008cheap}, the subjects are aware of the `implicit annotation task' but 
they are not paid. 
From this data, we built a document-by-emotion matrix $M_{DE}$, providing the voting percentages for each document 
in the eight affective dimensions available in Rappler. An excerpt is provided in Table \ref{tab:Document-by-Emotion-Matrix}.


The idea of using documents annotated with emotions is not new~\cite{strapparava2008learning,mishne2005experiments,bellegarda2010emotion}, but these works had the limitation of 
providing a single emotion label per document, rather than a score for each emotion, and, moreover, the annotation was performed by the author of the document alone.

Table \ref{tab:percentage-votes} reports the average percentage of votes for each emotion on the whole corpus: \textsc{happiness} has a far higher percentage of votes (at least three times). There 
are several possible explanations, out of the scope of the present paper, for this bias: (i) it is due to cultural characteristics of the audience 
(ii)
the bias is in the dataset itself, being formed mainly by `positive' news; (iii) it is a psychological phenomenon due to the fact that people tend to express more positive moods on 
social networks~\cite{querciamood,vittengl1998time,de2012not}. In any case, the predominance of happy mood has been found in other datasets, for instance \texttt{LiveJournal.com} posts \cite{strapparava2008learning}.
In the following section we will discuss how we handled this problem.

\begin{table} [!htb] 	
	\begin{center} 	 	
		{\footnotesize 		
			\begin{tabular}{lr|lrrr} 		
				\hline 
 EMOTION & Votes$_{\mu}$ & EMOTION & Votes$_{\mu}$\\
 \hline 
AFRAID & 0.04 & DONT\_CARE & 0.05 \\
AMUSED & 0.10 & HAPPY & 0.32 \\
ANGRY & 0.10 & INSPIRED & 0.10 \\
ANNOYED & 0.06 & SAD & 0.11 \\
\hline 
			\end{tabular} 		
		} 		 	
	\end{center}	 	
	\setlength{\belowcaptionskip}{-0.1cm} 	
	\caption{Average percentages of votes.} 	
	\label{tab:percentage-votes} 
\end{table}

\begin{table*} [!htb] 	
	\begin{center} 	 	
		{\footnotesize 		
			\begin{tabular}{l|rrrrrrrrr} 		
				\hline 
Word & AFRAID & AMUSED & ANGRY & ANNOYED & DONT\_CARE & HAPPY & INSPIRED & SAD \\
\hline
awe\#n & 0.08 & 0.12 & 0.04 & 0.11 & 0.07 & 0.15 & \textbf{0.38} & 0.05 \\
comical\#a & 0.02 & \textbf{0.51} & 0.04 & 0.05 & 0.12 & 0.17 & 0.03 & 0.06 \\
crime\#n & 0.11 & 0.10 & \textbf{0.23} & 0.15 & 0.07 & 0.09 & 0.09 & 0.15 \\
criminal\#a & 0.12 & 0.10 & \textbf{0.25} & 0.14 & 0.10 & 0.11 & 0.07 & 0.11 \\
dead\#a & 0.17 & 0.07 & 0.17 & 0.07 & 0.07 & 0.05 & 0.05 & \textbf{0.35} \\
funny\#a & 0.04 & \textbf{0.29} & 0.04 & 0.11 & 0.16 & 0.13 & 0.15 & 0.08 \\
future\#n & 0.09 & 0.12 & 0.09 & 0.12 & 0.13 & 0.13 & \textbf{0.21} & 0.10 \\
game\#n & 0.06 & 0.15 & 0.06 & 0.08 & 0.15 & \textbf{0.23} & 0.15 & 0.12 \\
kill\#v & \textbf{0.23} & 0.06 & \textbf{0.21} & 0.07 & 0.05 & 0.06 & 0.05 & \textbf{0.27} \\
rapist\#n & 0.02 & 0.07 & \textbf{0.46} & 0.07 & 0.08 & 0.16 & 0.03 & 0.12 \\
sad\#a & 0.06 & 0.12 & 0.09 & 0.14 & 0.13 & 0.07 & 0.15 & \textbf{0.24} \\
warning\#n & \textbf{0.44} & 0.06 & 0.09 & 0.09 & 0.06 & 0.06 & 0.04 & 0.16 \\
\hline
\end{tabular} 		
		} 	
				 	
	\end{center}	 	
	\setlength{\belowcaptionskip}{-0.1cm} 	
	\caption{An excerpt of the Word-by-Emotion Matrix ($M_{WE}$) using normalized frequencies ($nf$). Emotions weighting more than 20\% in a word are highlighted for readability purposes.} 	
	\label{tab:word-emotion} 
\end{table*}

\section{Emotion Lexicon Creation}

As a next step we built a word-by-emotion matrix starting from $M_{DE}$ using an approach based on compositional semantics. 
To do so, we first lemmatized and PoS tagged all the documents (where PoS can be adj., nouns, verbs, adv.) and kept only those \texttt{lemma\#PoS} present also in WordNet, similar to SWN-prior and WordNetAffect resources, to which we want to align.
We then computed the term-by-document matrices using raw frequencies, normalized 
frequencies, and tf-idf ($M_{WD,f}$, $M_{WD,nf}$ and $M_{WD,tfidf}$ respectively), so to test which of the three weights is better. 
After that, we applied matrix multiplication between the document-by-emotion and word-by-document matrices ($M_{DE} \cdot M_{WD}$) to obtain a (raw) word-by-emotion matrix $M_{WE}$. This method allows us to `merge' words with emotions by summing the products of the weight of a word with the weight of the emotions in each document. 

Finally, we transformed $M_{WE}$ 
by first applying 
normalization column-wise
(so to eliminate the over representation for happiness as discussed in Section \ref{DS}) 
and then scaling the data row-wise so to sum up to one. 
An excerpt of the final Matrix $M_{WE}$ is presented in Table \ref{tab:word-emotion}, and it can be interpreted as a list of 
words with scores that represent how much 
weight a given word has in the affective dimensions we consider. So, for example, \texttt{awe\#n} has a predominant weight in \textsc{inspired} (0.38), \texttt{comical\#a} has a predominant weight in \textsc{amused} (0.51), while \texttt{kill\#v} has a predominant weight in \textsc{afraid}, \textsc{angry} and \textsc{sad} (0.23, 0.21 and 0.27 respectively). This matrix, that we call \texttt{DepecheMood}\footnote{
In French, `depeche' means dispatch/news.}, represents our emotion lexicon, it contains 37k entries and is freely available for research purposes at http://git.io/MqyoIg.  


\section{Experiments}


To evaluate the performance we can obtain with our lexicon, we use the public dataset provided for the SemEval 2007 task on `Affective Text'~\cite{strapparava2007semeval}. The task was focused on emotion recognition in one thousand news headlines, 
both in regression and classification settings. Headlines typically consist of a few words and are often written with the 
intention to `provoke' emotions so to attract the readers' attention. An example of headline from the dataset is the following: ``\emph{Iraq car 
bombings kill 22 People, wound more than 60}". For the regression task the values provided are: \texttt{{\small $<$anger(0.32),disgust(0.27),fear(0.84), joy(0.0),sadness(0.95),surprise(0.20)$>$}} 
while for the classification task the labels provided are \texttt{\{FEAR,SADNESS\}}.
This  dataset is  of interest to us since the `compositional' problem is less prominent given the simplified 
syntax of news headlines, containing, for example, fewer adverbs (like negations or intensifiers) than normal sentences~\cite{turchi2012onts}. 
Furthermore, this is to our 
knowledge the only dataset available providing numerical scores for emotions. Finally, this dataset was meant for unsupervised approaches (just a small trial sample was 
provided), so to avoid simple text categorization approaches. 

As the affective dimensions present in the test set -- based on the six basic emotions model \cite{Ekman1971} -- do not exactly
 match with the ones provided by Rappler's Mood Meter, we first define a mapping between the two when possible, see
 Table~\ref{tab:mapping}. Then, we proceed to transform the test headlines to the \texttt{lemma\#PoS} format.
 
 \begin{table} [h] 	
	\begin{center} 	 	
		{\footnotesize 		
			\begin{tabular}{ll|ll} 						
			\hline 
SemEval  & Rappler  & SemEval  & Rappler \\
			\hline 
FEAR & AFRAID & \textbf{SURPRISE} & \textbf{INSPIRED} \\
ANGER & ANGRY & - & ANNOYED\\
JOY & HAPPY & - & AMUSED\\
SADNESS & SAD & - & DON'T CARE\\
\hline

			\hline 		
			\end{tabular} 		
		} 		 	
	\end{center}	 	
	\setlength{\belowcaptionskip}{-0.1cm} 	
	\caption{Mapping of Rappler labels on Semeval2007. In bold, cases of suboptimal mapping.}
	\label{tab:mapping} 
\end{table}  

Only one test headline contained exclusively words not present in \texttt{DepecheMood}, further indicating the high-coverage nature of our resource. In Table 
\ref{tab:coverage} we report the coverage of some Sentiment and Emotion Lexica of different sizes on the same dataset. 
Similar to Warriner et al. (2013), we 
observe that even if the number of entries of our lexicon is far lower than SWN-prior approaches, the fact that we extracted and annotated words from documents grants a high 
coverage of language use. 

\begin{table} [h] 	
	\begin{center} 	 	
		{\footnotesize 		
			\begin{tabular}{l|lrr} 						
			\hline 

\hline			 
\multirow{3}{*}{\parbox[t]{1.6cm}{Sentiment Lexica}}& ANEW &1k entries & 0.10 \\
&Warriner et. al & 13k entries & 0.51 \\
&SWN-prior & 155k entries & \textbf{0.67} \\				  
\hline
\multirow{2}{*}{\parbox[t]{1.5cm}{Emotion Lexica}}&WNAffect &1k entries & 0.12 \\
&DepecheMood &37k entries & \textbf{0.64} \\
 \hline
			\end{tabular} 		
		} 		 	
	\end{center}	 	
	\setlength{\belowcaptionskip}{-0.1cm} 	
	\caption{Statistics on words coverage per headline.} 	
	\label{tab:coverage} 
\end{table} 

Since our primary goal is to assess the quality of  \texttt{DepecheMood} 
we first focus on the regression task. 
We do so by using a very na\"{\i}ve approach, similar to ``WordNetAffect presence" discussed in \cite{strapparava2008learning}: for each  headline,
we simply compute a value, for any affective dimension, by averaging the corresponding affective scores --obtained from \texttt{DepecheMood}\-- of all \texttt{lemma\#PoS} present in the headline.


In Table \ref{tab:SemEval2007} we report the results obtained using the three versions of our resource (Pearson correlation), along with the best performance on each emotion of 
other systems\footnote{Systems participating in the `Affective Text' task plus the approaches in ~\cite{strapparava2008learning}. Other supervised approaches in the classification task \cite{mohammad2012emotional,bellegarda2010emotion,chaffar2011using}, reporting only overall performances, are not considered.} ($best_{se}$); the last column contains the upper bound of inter-annotator agreement.
For all the 5 emotions we improve over the best performing 
systems (\texttt{DISGUST} has no alignment with our labels and was discarded).

Interestingly, even using a sub-optimal alignment for \texttt{SURPRISE} we still manage to outperform other systems. Considering the na\"{\i}ve approach we used, we can 
reasonably conclude that the quality and coverage of our resource are the reason of such results, and that adopting more complex approaches (i.e. compositionality) can possibly further improve 
performances in text-based emotion recognition.

\begin{table} [htb!] 	
	\begin{center} 	 	
		{\footnotesize 		
			\begin{tabular}{l|rrr|r|r} 		
				\hline 
& \multicolumn{3}{c|}{$DepecheMood$} & $best_{se}$ & upper\\
 & \emph{f} & \emph{nf} & \emph{tfidf} &  &  \\
FEAR & \textbf{0.56} & 0.54 & 0.53 & 0.45 & 0.64 \\
ANGER & 0.36 & \textbf{0.38} & 0.36 & 0.32 & 0.50 \\
SURPRISE* & \textbf{0.25} & 0.21 & 0.24 & 0.16 & 0.36 \\
JOY & 0.39 & \textbf{0.40} & 0.39 & 0.26 & 0.60 \\
SADNESS & \textbf{0.48} & 0.47 & 0.46 & 0.41 & 0.68 \\

\hline 		
			\end{tabular} 		
		} 		 	
	\end{center}	 	
	\setlength{\belowcaptionskip}{-0.1cm} 	
	\caption{Regression results -- Pearson's correlation} 	
	\label{tab:SemEval2007}  
\end{table}  

As a final test, we evaluate our resource in the classification task. 
The  na\"{\i}ve approach used in this case consists in mapping the average of the scores of 
all words in the headline to a binary decision with fixed threshold at 0.5 for each emotion (after min-max normalization on all test headlines scores). 
In Table~\ref{tab:SemEval2007_class} we report the results (F1 measure) of our 
approach along with the best performance of other systems on each emotion ($best_{se}$), as in the previous case. 
For 3 emotions out of 5 we improve over the best performing systems, for one emotion we obtain the same results, and for one emotion we do not outperform other systems. In this case the difference 
in performances among the various ways of representing the word-by-document matrix is more prominent: normalized frequencies  ($nf$) provide the best results. 

\begin{table} [htb!] 	
	\begin{center} 	 	
		{\footnotesize 		
			\begin{tabular}{l|rrr|r} 		
				\hline 
& \multicolumn{3}{c|}{$DepecheMood$} & $best_{se}$ \\
 & \emph{f} & \emph{nf} & \emph{tfidf} & \\
FEAR & 0.25 & \textbf{0.32} & 0.31 & 0.23 \\
ANGER & 0.00 & 0.00 & 0.00 & \textbf{0.17} \\
SURPRISE* & 0.13 & \textbf{0.16} & 0.09 & 0.15 \\
JOY & 0.22 & 0.30 & \textbf{0.32} & \textbf{0.32} \\
SADNESS & 0.36 & \textbf{0.40} & 0.38 & 0.30 \\
\hline 		
			\end{tabular} 		
		} 		 	
	\end{center}	 	
	\setlength{\belowcaptionskip}{-0.1cm} 	
	\caption{Classification results -- F1 measures} 	
	\label{tab:SemEval2007_class} 
\end{table} 

\section{Conclusions}
We presented 
\texttt{DepecheMood}, an emotion lexicon built in a novel and totally automated way by harvesting crowd-sourced affective annotation from a social news network. Our experimental results indicate 
high-coverage and high-precision of the lexicon, showing significant improvements over state-of-the-art unsupervised approaches even when using the resource with
very na\"{\i}ve classification and regression strategies.
We believe that the wealth of information provided by social media can be harnessed to build models and resources for emotion recognition from text, going  a step beyond 
sentiment analysis.
Our future work will include testing Singular Value Decomposition on the word-by-document matrices, allowing to propagate emotions values for a document to similar 
words non present in the document itself, and the study of perceived mood effects on virality indices and readers engagement by exploiting tweets, likes, reshares and comments.\makeatletter{\renewcommand*\@makefnmark{}
\footnotetext{This work has been partially supported by the Trento RISE PerTe project.}
\makeatother} 



\bibliographystyle{acl}
\bibliography{Persuasive}

\end{document}